\documentclass[conference]{IEEEtran}
\IEEEoverridecommandlockouts
\usepackage{cite}
\usepackage{times}
\usepackage{listings}
\usepackage{tabularray}
\usepackage{soul}
\usepackage{url}
\usepackage{amsmath,amssymb,amsfonts}
\usepackage{algorithmic}
\usepackage{listings}
\usepackage{graphicx}
\usepackage{hyperref}
\usepackage{textcomp}
\usepackage{xcolor}
\usepackage{nicematrix}
\usepackage{array}
\usepackage{multirow}
\usepackage{ragged2e}
\usepackage{mathtools}
\usepackage{subcaption}
\usepackage[labelformat=parens,labelsep=quad, skip=3pt]{caption}
\usepackage{graphicx}
\usepackage{todonotes} 

\newtheorem{example}{Example}[section]
\def\BibTeX{{\rm B\kern-.05em{\sc i\kern-.025em b}\kern-.08em
    T\kern-.1667em\lower.7ex\hbox{E}\kern-.125emX}}

    \makeatletter
    \newcommand{\linebreakand}{%
      \end{@IEEEauthorhalign}
      \hfill\mbox{}\par
      \mbox{}\hfill\begin{@IEEEauthorhalign}
    }
    \makeatother

\begin{document}

\title{Knowledge Enhanced Graph Neural Networks
}
\author{
\IEEEauthorblockN{1\textsuperscript{st} Luisa Werner}
\IEEEauthorblockA{
\textit{Université Grenoble Alpes, INRIA}\\
Grenoble, France\\
luisa.werner@inria.fr}
\and
\IEEEauthorblockN{2\textsuperscript{nd} Nabil Laya\"ida}
\IEEEauthorblockA{
\textit{INRIA}\\
Grenoble, France\\
nabil.layaida@inria.fr}
\and
\IEEEauthorblockN{3\textsuperscript{rd} Pierre Genevès}
\IEEEauthorblockA{\textit{CNRS, INRIA}\\
Grenoble, France\\
pierre.geneves@inria.fr}
\and
\centering
\IEEEauthorblockN{4\textsuperscript{th} Sarah Chlyah}
\IEEEauthorblockA{\textit{INRIA}\\
Grenoble, France\\
sarah.chlyah@inria.fr}
}

\maketitle

\begin{abstract}
  Graph data is omnipresent and has a wide variety of applications, such as in natural science, social networks, or the semantic web. 
  However, while being rich in information, graphs are often noisy and incomplete.
  As a result, graph completion tasks, such as node classification or link prediction, have gained attention. On one hand, neural methods, such as graph neural networks, have proven to be robust tools for learning rich representations of noisy graphs. On the other hand, symbolic methods enable exact reasoning on graphs.
  We propose Knowledge Enhanced Graph Neural Networks (KeGNN), a neuro-symbolic framework for graph completion that combines both paradigms as it allows for the integration of prior knowledge into a graph neural network model.
  Essentially, KeGNN consists of a graph neural network as a base upon which knowledge enhancement layers are stacked with the goal of refining predictions with respect to prior knowledge.
  We instantiate KeGNN in conjunction with two well-known graph neural networks, Graph Convolutional Networks and Graph Attention Networks, and evaluate KeGNN on multiple benchmark datasets for node classification.
\end{abstract}


\section{Introduction}
Graphs are ubiquitous across diverse real-world applications such as e-commerce 
\cite{gnn_ecommerce}
, natural science 
\cite{pmlr-v80-sanchez-gonzalez18a}
or social networks 
\cite{Wu_Lian_Xu_Wu_Chen_2020}.
Graphs connect nodes by edges and allow to enrich them with features. 
This makes them a versatile and powerful data structure that encodes relational information.
As graphs are often derived from noisy data, incompleteness and errors are common issues.
Consequently, graph completion tasks such as node classification or link prediction have become increasingly important. 
These tasks are approached from different directions. 
In the field of deep learning, research on graph neural networks (GNNs) has gained momentum.
Numerous models have been proposed for various graph topologies and applications 
\cite{heterogeneous_gnn, dl_on_graphs, gnn_survey, gnn_kg}. 
The key strength of GNNs is to find meaningful representations of noisy graph data, that can be used to improve prediction tasks \cite{GNNBook2022}.
Despite this advantage, as a subcategory of deep learning methods, GNNs are criticized for their limited interpretability and large data consumption
\cite{neuro-symbolic_survey}. 
Alongside, the research field of symbolic AI addresses the above-mentioned tasks.
In symbolic AI, solutions are found by performing logic-like reasoning steps that are exact, interpretable and data-efficient \cite{survey_nesy}.
For large graphs, however, symbolic methods are often computationally expensive or even infeasible.
Since techniques from deep learning and from symbolic AI have complementary pros and cons, the field of neuro-symbolic AI aims to combine both paradigms. 
Neuro-symbolic AI not only paves the way towards the application of AI to learning with limited data, but also allows for jointly using symbolic information (in the form of logical rules) and sub-symbolic information (in the form of real-valued data).
This helps to overcome the black-box nature of deep learning methods and to improve interpretability through symbolic representations 
\cite{GARNELO201917, rnm, neuro-symbolic_survey}. 

In this work, we present the neuro-symbolic approach \textbf{K}nowledge \textbf{e}nhanced \textbf{G}raph \textbf{N}eural \textbf{N}etworks (KeGNN) to conduct node classification given 
graph data and a set of prior knowledge.
In KeGNN, knowledge enhancement layers \cite{kenn} are stacked on top of a GNN and adjust its predictions in order to increase the satisfaction of some prior knowledge.
In addition to the parameters of the GNN, the knowledge enhancement layers contain learnable clause weights that reflect the impact of the prior knowledge on the predictions.
Both components form an end-to-end differentiable model.
KeGNN can be seen as a variant of knowledge enhanced neural networks (KENN), which stack knowledge enhancement layers onto a multi-layer perceptron (MLP) and have been proven successful in semantic point cloud segmentation, image segmentation and multi-label classification \cite{relational_kenn, kenn_pointclouds,kenn_original}. 
However, relational information in sparse graphs can only be introduced through the logical clauses with binary predicates in the knowledge enhancement layer and not at base neural network level.
In contrast, KeGNN is based on GNNs that process the graph structure, which makes both the neural and symbolic components sufficiently powerful to exploit the graph structure.
In this work, we instantiate KeGNN in conjunction with two well-known GNNs: Graph Attention Networks \cite{gat} and Graph Convolutional Networks \cite{gcn}. We apply KeGNN to the benchmark datasets for node classification Cora, Citeseer, PubMed \cite{planetoid_cite} and Flickr \cite{graph_saint}.

\section{Method: KeGNN}
KeGNN is a neuro-symbolic approach that can be applied to node classification tasks with the capacity of handling graph structure at the base neural network level. 
The model takes two types of input: (1) real-valued graph data and (2) prior knowledge expressed in first-order logic. 
\subsection{Graph-structured Data}
\label{sec:graphstructured_data}
\begin{figure*}[h] 
  \centering
  \centering\includegraphics[width=12cm]{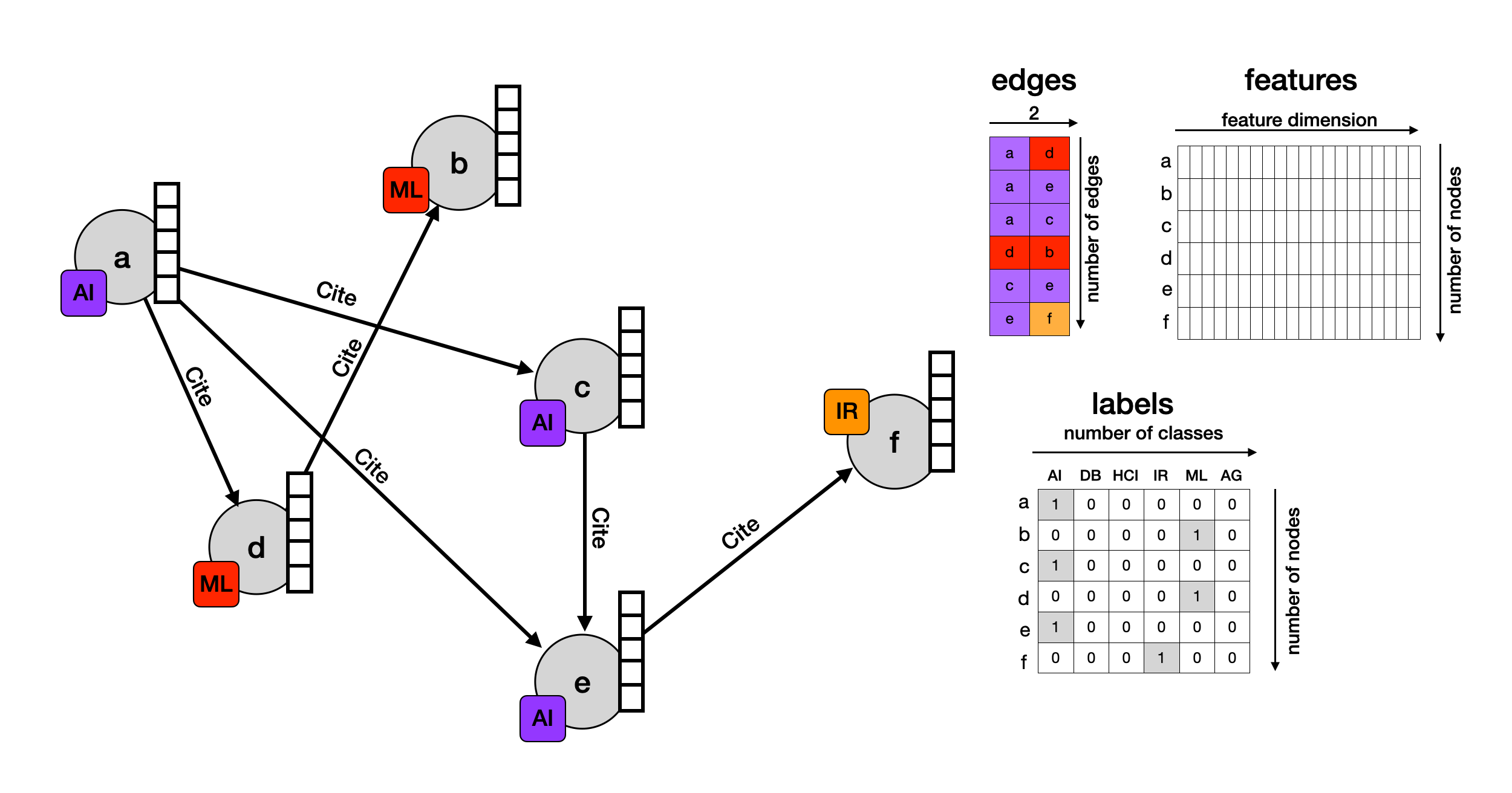}
  \caption{
      Example extract of the Citeseer citation graph. 
  \label{fig:graph_structure}}
\end{figure*}%
A Graph $\mathbf{G}=(\mathbf{N}, \mathbf{E})$ consists of a set of $n$ nodes $\mathbf{N}$ and a set of $k$ edges $\mathbf{E}$ where each edge of the form $(v_i, v_j)$ connects two nodes $v_i \in \mathbf{N}$ and $v_j \in \mathbf{N}$.
The neighborhood $\mathcal{N}(v_i)$ describes the set of first-order neighbors of $v_i$. 
For an \emph{attributed} and \emph{labelled} graph, nodes are enriched with features and labels.
Each node has a feature vector $\mathbf{x} \in \mathbb{R}^{d}$ of dimension $d$ and a label vector $\mathbf{y} \in \mathbb{R}^m$. 
The label vector $\mathbf{y}$ contains one-hot encoded ground truth labels for $m$ classes.
In matrix notation, the features and labels of the entire graph are described as $\mathbf{X} \in \mathbb{R}^{n \times d}$ and $\mathbf{Y} \in \mathbb{R}^{n \times m}$.
A graph is \emph{typed} if type functions $f_{\mathbf{E}}$ and $f_{\mathbf{N}}$ assign edge types and node types to the edges and nodes, respectively. 
A graph with constant type functions (that assign the same edge and node type to all edges and nodes) is called \emph{homogeneous}, whereas for \emph{heterogeneous} graphs, nodes and edges may have different types \cite{dl_on_graphs}.

\begin{example}
    \label{ex:graph_cit}
    A Citation Graph $\mathbf{G}_{\mathrm{Cit}}$ consists of documents and citations.
    Figure~\ref{fig:graph_structure} shows an extract of the Citeseer citation graph that is used as example to guide through this paper. 
    The documents are represented by a set of nodes $\mathbf{N}_{\mathrm{Cit}}$ and citations by a set of edges $\mathbf{E}_{\mathrm{Cit}}$.
    Documents can be attributed with features $\mathbf{X}_{\mathrm{Cit}}$ that describe their content as Word2Vec \cite{word2vec} vectors.
    Each node is labelled with one of the six topic categories \{AI, DB, HCI, IR, ML, AG\}\footnote{The classes are abbreviations for the categories \emph{Artificial Intelligence, Databases, Human-Computer Interaction, Information Retrieval, Machine Learning and Agents}.} 
    that are encoded in $\mathbf{Y}_{\mathrm{Cit}}$.
    Since all nodes (documents) and edges (citations) have the same type, $\mathbf{G}_{\mathrm{Cit}}$ is homogeneous. 
    \end{example}
    \subsection{Prior Knowledge} 
\label{sec:prior_knowledge}
Some prior knowledge $\mathcal{K}$ is provided to KeGNN.
It can be described as a set of $\ell$ logical clauses expressed in the logical language $\mathcal{L}$ that is defined as sets of constants $\mathcal{C}$, variables $\mathcal{X}$ and predicates $\mathcal{P}$.
Predicates have an arity $r$ of one (unary) or two (binary): $\mathcal{P} = \mathcal{P}_U$ $\cup$ $\mathcal{P}_B$.
Predicates of arity $r>2 $ are not considered in this work.
Unary predicates express properties, whereas binary predicates express relations. 
$\mathcal{L}$ supports negation ($\neg$) and disjunction ($\lor$).
Each clause $\varphi \in \mathcal{K} = \{\varphi_1, \hdots , \varphi_{\ell}\}$ can be formulated as a disjunction of (possibly negated) atoms ${\bigvee_{j=1}^{q} o_{j}}$ with $q$ atoms $\{o_1, \hdots, o_q\}$.
Since the prior knowledge is general, all clauses are assumed to be universally quantified.
Clauses can be \emph{grounded} by assigning constants to the free variables.
A grounded clause is denoted as $\varphi[x_1,x_2,...|c_1, c_2,...]$ with variables $x_i \in \mathcal{X}$ and constants $c_i \in \mathcal{C}$.
The set of all grounded clauses in a graph is $\mathcal{G}(\mathcal{K}, \mathcal{C})$. 

\begin{example}
\label{ex:prior_knowledge_cit} 
The graph $\mathbf{G}_{\mathrm{Cit}}$ in Figur~\ref{fig:graph_structure} can be expressed in $\mathcal{L}$.
Nodes are represented by a set of constants $\mathcal{C}=\{a, b, \hdots,f\}$.
Node labels are expressed as a set of unary predicates $\mathcal{P}_U=\{\mathrm{AI, DB, \hdots, AG}\}$ and edges as a set of binary predicates $\mathcal{P}_B = \{\mathrm{Cite}\}$.
$\mathcal{L}$ has a set of variables $\mathcal{X} = \{x,y\}$. 
The atom $\mathrm{AI(x)}$, for example, expresses the membership of $x$ to the class $\mathrm{AI}$ and $\mathrm{Cite(x,y)}$ expresses the existence of a citation between $x$ and $y$.
Some prior knowledge $\mathcal{K}$ can be written as a set of $\ell=6$ disjunctive clauses in $\mathcal{L}$. 
Here, the assumption is denoted that two papers that cite each other have the same document class:
\begin{align*}
  \varphi_{\mathrm{AI}}: \forall xy\neg \mathrm{AI(x)} \lor \neg \mathrm{Cite(x,y)} \lor \mathrm{AI(y)}\\
  \varphi_{\mathrm{DB}}: \forall xy \neg \mathrm{DB(x)} \lor \neg \mathrm{Cite(x,y)} \lor \mathrm{DB(y)}\\
  \hdots \quad \quad \quad \quad \quad \quad \quad 
\end{align*}
\normalsize
The atoms are grounded by replacing the variables $x$ and $y$ with the constants $\{a, b, \hdots f\}$ to obtain the sets of unary groundings $\{\mathrm{AI(a), ML(b),} \hdots \mathrm{ , IR(f)}\}$ and binary groundings $\{\mathrm{Cite(a,d), Cite(a,e),} \hdots, \mathrm{Cite(a,f)}\}$.
Assuming a closed world and exclusive classes, other facts could be derived, such as $\{\neg \mathrm{DB(a)}, \neg \mathrm{IR(a)}, \hdots, \neg \mathrm{Cite(a,b)}\}$.
For the sake of simplicity, these are omitted here.  
\end{example}

\subsection{Node Classification}
Node classification is a subtask of knowledge graph completion on a graph $\mathbf{G}$ with the objective to assign classes to nodes where they are unknown. 
This task is accomplished given node features $\mathbf{X}$, edges $\mathbf{E}$ and some prior knowledge $\mathcal{K}$ encoded as a set of clauses in $\mathcal{L}$.
A predictive model is trained on a subset of the graph $\mathbf{G}_{\text{train}}$ with ground truth labels $\mathbf{Y}_{\text{train}}$ and validated on a test set $\mathbf{G}_{\text{test}}$ for which the ground truth labels are compared to the predictions in order to assess the predictive performance.
Node classification can be studied in a \emph{transductive} or \emph{inductive} setting. 
In a transductive setting, the entire graph is available for training, but the true labels of the test nodes are masked. 
In an inductive setting, only the nodes in the training set and the edges connecting them are available, making it more challenging to classify unseen nodes.

\subsection{Fuzzy Semantics}
\label{sec:fuzzy_semantics}
Let us consider an attributed and labelled graph $\mathbf{G}$ and the prior knowledge  $\mathcal{K}$.
While $\mathcal{K}$ can be defined in the logic language $\mathcal{L}$,
the neural component in KeGNN relies on continuous and differentiable representations. 
To interpret Boolean logic in the real-valued domain, KeGNN uses fuzzy logic \cite{fuzzy_logic}, which maps Boolean truth values to the continuous interval $[0,1] \subset \mathbb{R}$.
A constant in $\mathcal{C}$ is interpreted as a real-valued feature vector $\mathbf{x} \in \mathbb{R}^{d}$. 
A predicate $P \in \mathcal{P}$ with arity $r$ is interpreted as a function $f_{P}:  \mathbb{R}^{r \times d} \mapsto [0,1]$ that takes $r$ feature vectors as input and returns a truth value.

\begin{example}
  In the example, a unary predicate $P_U \in \mathcal{P}_U = \{\mathrm{AI, DB, \hdots}\}$ is interpreted as a function $f_{P_U}:\mathbb{R}^{d} \mapsto [0,1]$ that takes a feature vector $\mathbf{x}$ and returns a truth value indicating whether the node belongs to the class encoded as $P_U$. 
  The binary predicate $\mathrm{Cite} \in \mathcal{P}_B$ is interpreted as the function
 \begin{equation*}
    f_{\mathrm{Cite}}(v_i, v_j) = 
    \begin{cases}
      1, & \text{if } (v_i, v_j) \in \mathbf{E}_{\mathrm{Cit}} \\
      0, & \text{else.}
    \end{cases}
  \end{equation*}
 $f_{\mathrm{Cite}}$ returns the truth value 1 if there is an edge between two nodes $v_i$ and $v_j$ in $\mathbf{G}_{\mathrm{Cit}}$ and 0 otherwise.
\end{example}

T-conorm functions $\bot: [0,1] \times [0,1] \mapsto [0,1]$ \cite{klement2013triangular} take real-valued truth values of two literals\footnote{A literal is a (possibly negated) grounded atom, e.g. $\mathrm{AI(a)}$} and define the truth value of their disjunction.
The G\"odel t-conorm function for two truth values $\mathbf{t}_i, \mathbf{t}_j$ is defined as
\begin{equation*}
  \bot(\mathbf{t}_i, \mathbf{t}_j) \mapsto \operatorname{max}(\mathbf{t}_i, \mathbf{t}_j).
\end{equation*}
To obtain the truth value of a clause $\varphi: o_1\lor ... \lor o_q$, the function $\bot$ is extended to a vector $\mathbf{t}$ of $q$ truth values: $\bot(\mathbf{t}_1, \mathbf{t}_2, ..., \mathbf{t}_q) = \bot(\mathbf{t}_1,\bot(\mathbf{t}_2... \bot(\mathbf{t}_{q-1}, \mathbf{t}_q)))$. Fuzzy negation over truth values is defined as $\mathbf{t} \mapsto 1-\mathbf{t}$ \cite{fuzzy_logic}.

\begin{example}
Given the clause $\varphi_{\mathrm{AI}}:$ $\forall xy$ $\neg \mathrm{AI(x)} \lor \neg \mathrm{Cite(x,y)} \lor \mathrm{AI(y)}$ and its grounding $\varphi_{\mathrm{AI}}[x,y|a,b ]: \mathrm{AI(a)} \lor \neg \mathrm{Cite(a,b)} \lor \mathrm{AI(b)}$ to the constants $a$ and $b$ and truth values for the grounded predicates $\mathrm{AI(a)}=\mathbf{t}_1$, $\mathrm{AI(b)}=\mathbf{t}_2$ and $\mathrm{Cite(a,b)}=\mathbf{t}_3$, the truth value of $\varphi_{\mathrm{AI}}[x,y|a,b]$ is
$\mathrm{max}\{\mathrm{max}\{(1-\mathbf{t}_1), (1-\mathbf{t}_3)\}, \mathbf{t}_2\}$. 

\end{example}
\subsection{Model Architecture}
\begin{figure*}[h]
  \centering
  \includegraphics[scale=0.3]
  {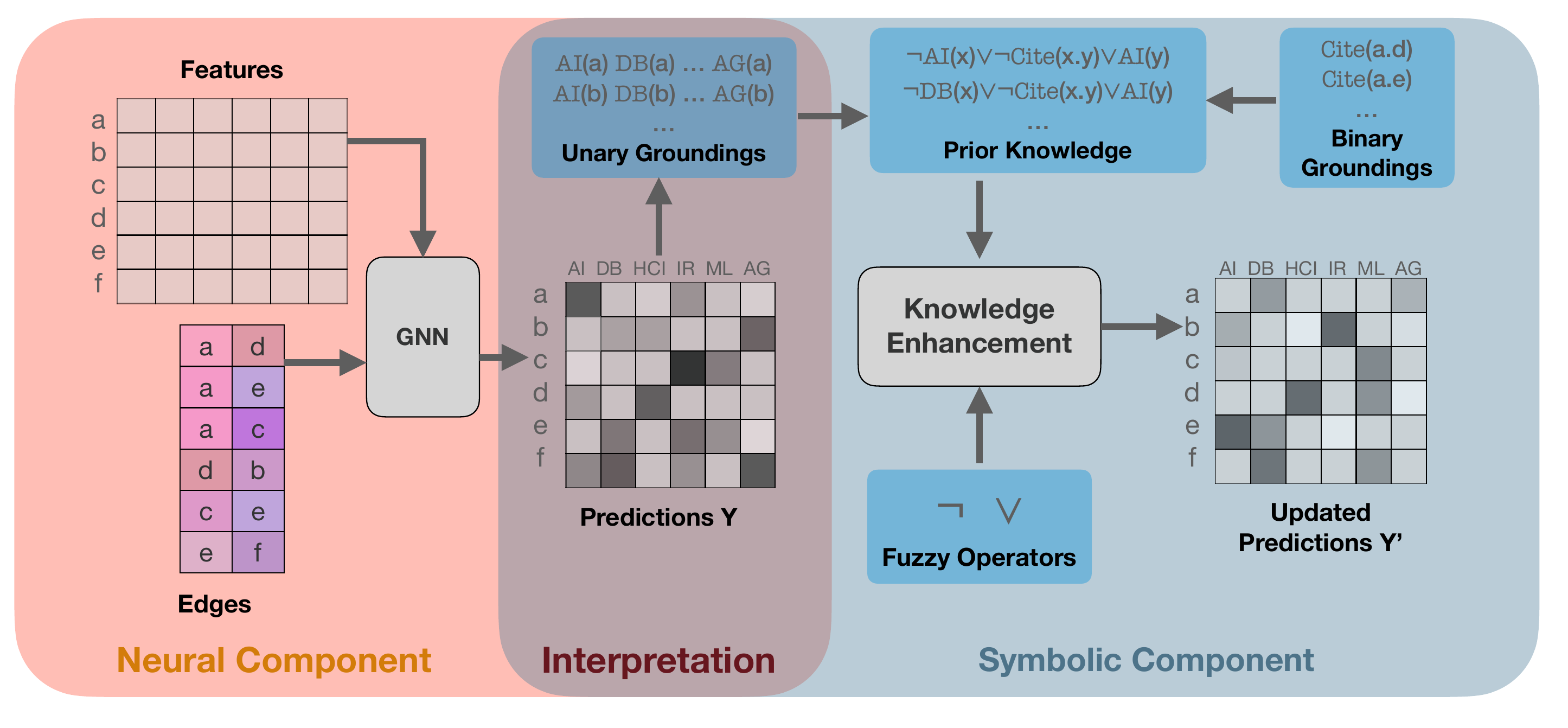}
  \caption{
      Overview of KeGNN. 
  \label{fig:kegnn}}
  \end{figure*}
The way KeGNN computes the final predictions can be divided in two stages. 
First, a GNN predicts the node classes given the features and the edges. 
Subsequently, the knowledge enhancement layers use the predictions as truth values for the grounded unary predicates and update them with respect to the knowledge.
An overview of KeGNN is given in Figure~\ref{fig:kegnn}.
%
\subsubsection{Neural Component}
The role of the GNN in the neural component is to exploit feature information in the graph structure.
The key strength of a GNN is to enrich node representations with graph structure by nesting 
$k$ message passing layers \cite{GNNBook2022}. 
Per layer, the representations of neighboring nodes are aggregated and combined to obtain updated representations. 
The node representation $v_i^{k+1}$ in the $k$-th message passing layer is 
\begin{equation*}
  v_i^{k+1} = 
    \operatorname{combine} \bigl( v_i^k, \operatorname{aggregate} \bigl(\{v_j^k | v_j^k \in \mathcal{N}(v_i)\}\bigr) \bigr).
\end{equation*}
The layers contain learnable parameters that are optimized with backpropagation. 
In this work, we consider two well-known GNNs as components for KeGNN: Graph Convolutional Networks (GCN) \cite{gcn} and Graph Attention Networks (GAT) \cite{gat}.
While GCN considers the graph structure as given, GAT allows for assessing the importance of the neighbors with attention weights $\alpha_{ij}$ between node $v_i$ and node $v_j$.
In case of multi-head attention, the attention weights are calculated multiple times and concatenated which allows for capturing different aspects of the input data.
In KeGNN, the GNN implements the functions $f_{P_U}$ (see Section~\ref{sec:fuzzy_semantics}). 
In other words, the predictions are used as truth values for the grounded unary predicates in the symbolic component.
\subsubsection{Symbolic Component}

To refine the predictions of the GNN, one or more knowledge enhancement layers are stacked onto the GNN to update its predictions $\mathbf{Y}$ to $\mathbf{Y}'$. 
The goal is to increase the satisfaction of the prior knowledge.
The predictions $\mathbf{Y}$ of the GNN serve as input to the symbolic component where they are interpreted as fuzzy truth values for the unary grounded predicates $\mathbf{U}:=\mathbf{Y}$ with $\mathbf{U} \in \mathbb{R}^{n \times m}$.
Fuzzy truth values for the groundings of binary predicates are encoded as a matrix $\mathbf{B}$ where each row represents an edge $(v_i, v_j)$ and each column represents an edge type $e$.
In the context of node classification, the GNN returns only predictions for the node classes, while the edges are assumed to be given.
A binary grounded predicate is therefore set to truth value $1$ (true) if an edge between two nodes $v_i$ and $v_j$ exists:
\begin{equation*}
  \mathbf{B}_{[(v_i, v_j), e]} = 
  \begin{cases}
    1, & \text{if $(v_i, v_j)$ of type $e \in \mathbf{E}$} \\
    0, & \text{else.}
  \end{cases}
\end{equation*}

\begin{example}
In case of the beforementioned citation graph of Figure~\ref{fig:graph_structure}, $\mathbf{U}$ and $\mathbf{B}$ are defined as: 
\begin{equation*}
  \label{eq:unary_predicates}
  \mathbf{U} := 
  \begin{bmatrix}
      \mathrm{AI(a)}  & \hdots & \mathrm{AG(a)} \\
      \mathrm{AI(b)} & \hdots & \mathrm{AG(b)} \\
      \vdots & & \vdots \\
      \mathrm{AI(f)} & \hdots &  \mathrm{AG(f)}\\
  \end{bmatrix}
  \quad
  \mathbf{B} := 
  \begin{bmatrix}
      \mathrm{Cite(a,d)} \\
      \mathrm{Cite(a,e)} \\
      \mathrm{Cite(a,c)} \\
      \vdots \\
      \mathrm{Cite(c,e)} \\
      \mathrm{Cite(e,f)} \\
  \end{bmatrix}
\end{equation*}
\end{example}
To enhance the satisfaction of clauses that contain both unary and binary predicates, their groundings are joined into one matrix $\mathbf{M} \in \mathbb{R}^{k \times p}$ with $p = 2\cdot|\mathcal{P}_U| + |\mathcal{P}_B|$. 
$\mathbf{M}$ is computed by joining $\mathbf{U}$ and $\mathbf{B}$ so that each row of $\mathbf{M}$ represents an edge $(v_i, v_j)$.  
As a result, $\mathbf{M}$ contains all required grounded unary predicates for $v_i$ and $v_j$. 
\begin{example}
For the example citation graph, we obtain $\mathbf{M}$ as follows:

\includegraphics[trim=1.2cm 0 0 0cm, scale=0.34]{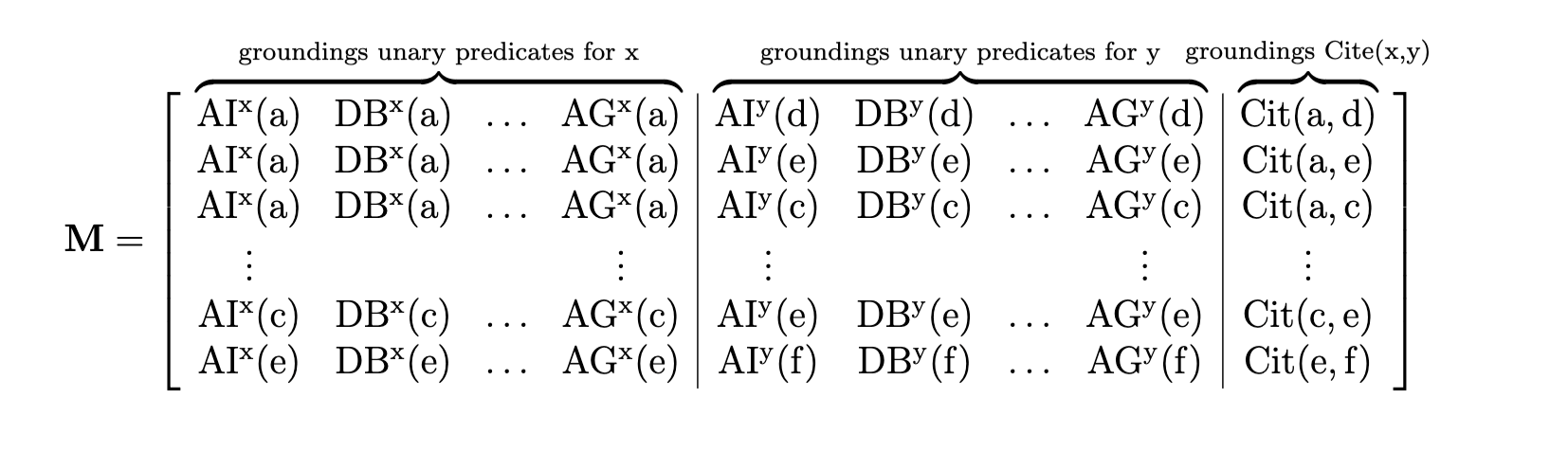}
\end{example}
A knowledge enhancement layer consists of multiple \emph{clause enhancers}.
A clause enhancer is instantiated for each clause $\varphi \in \mathcal{K}$.  
Its aim is to compute updates $\delta \mathbf{M}_{\varphi}$ for the groundings in $\mathbf{M}$ that increase the satisfaction of $\varphi$. 

First, fuzzy negation is applied to the columns of $\mathbf{M}$ that correspond to negated atoms in $\varphi$. Then $\delta \mathbf{M}_{\varphi}$ is computed by a \emph{t-conorm boost function} $\phi$ \cite{kenn}. 
This function $\phi: [0,1]^q \mapsto [0,1]^q$ takes $q$ truth values and returns changes to those truth values such that the satisfaction is increased: $\bot(\mathbf{t}) \le \bot(\mathbf{t} + \phi(\mathbf{t}))$.
\cite{kenn} propose the following differentiable t-conorm boost function 
$$\phi_{w_{\varphi}}(\mathbf{t})_i = w_{\varphi} \cdot \frac{e^{\mathbf{t}_i}}{\sum_{j=1}^q e^{\mathbf{t}_j}}.$$ 
The boost function $\phi_{w_{\varphi}}$ employs a clause weight $w_{\varphi}$ that is initialized in the beginning of the training and optimized during training
as a learnable parameter.
The updates for the groundings calculated by $\phi_{w_{\varphi}}$ are proportional to $w_{\varphi}$.
Therefore, $w_{\varphi}$ determines the magnitude of the update and thus reflects the impact of a clause.
The changes to the atoms that do not appear in a clause are set to zero. 
The boost function is applied row-wise to $\mathbf{M}$ as illustrated in the following example.  
\begin{example}
Given the clause $\varphi_{AI}: \forall xy \neg \mathrm{AI(x)} \lor \neg \mathrm{Cite(x,y)} \lor \mathrm{AI(y)}$ and the clause weight $w_{\mathrm{AI}}$, the changes for this clause are $\delta \mathbf{M}_{\varphi_{AI}} =$ 
\begin{equation*} 
  \footnotesize
 w_{\mathrm{AI}} \cdot
  \begin{bNiceArray}{ccccccc}[margin]
      \delta_{\neg \mathrm{AI^x(a)}} & 0 & \hdots &\delta_{\mathrm{AI^y(c)}} & 0 & \hdots &\delta_{\neg \mathrm{Cit(a,c)}}\\
      \delta_{\neg \mathrm{AI^x(a)}} & 0 & \hdots &\delta_{\mathrm{AI^y(e)}} & 0 & \hdots &\delta_{\neg \mathrm{Cit(e,a)}}\\
      \delta_{\neg \mathrm{AI^x(a)}} & 0 & \hdots &\delta_{\mathrm{AI^y(d)}} & 0 & \hdots &\delta_{\neg \mathrm{Cit(c,d)}}\\
      \vdots &  & & \vdots & \vdots \\
      \delta_{\neg \mathrm{AI^x(e)}} & 0 & \hdots&\delta_{\mathrm{AI^y(f)}} & 0 & \hdots &\delta_{\neg \mathrm{Cit(e,f)}}\\
      \CodeAfter
  \end{bNiceArray}
\end{equation*}
The values of $\delta \mathbf{M}_{\varphi_{AI}}$ are calculated by $\phi_{w_{\mathrm{AI}}}$, for example: 
\begin{equation*}
  \footnotesize
  \delta_{\neg \mathrm{AI^x(a)}} = \phi_{w_{\mathrm{AI}}}(\mathbf{z})_a = - \frac{e^{-\mathbf{z}_{AI(a)}}}{e^{-\mathbf{z}_{AI(a)}} + e^{-\mathbf{z}_{Cit(a,c)}} + e^{\mathbf{z}_{AI(c)}}}
\end{equation*}
\end{example}
Each clause enhancer computes updates $\delta \mathbf{M}_{\varphi}$ to increase the satisfaction of a clause independently. 
The updates of all clause enhancers are finally added, resulting in a matrix $\delta\mathbf{M} = \sum_{\varphi \in \mathcal{K}} \delta \mathbf{M}_\varphi$. 
To apply the updates to the initial predictions, $\delta\mathbf{M}$ has to be added to $\mathbf{Y}$. 
The updates in $\delta \mathbf{M}$ can not directly be applied to the predictions $\mathbf{Y}$ of the GNN.
Since the unary groundings $\mathbf{U}$ were joined with the binary groundings $\mathbf{B}$, multiple changes may be proposed for the same grounded unary atom. 
For example, for the grounded atom $\mathrm{AI(c)}$ the changes $\delta_{\neg{\mathrm{AI^y(c)}}}$ and $\delta_{\neg{\mathrm{AI^x(c)}}}$ are proposed, since $c$ appears in the grounded clauses $\varphi_{\mathrm{AI}}[x,y|a,c]$ and $\varphi_{\mathrm{AI}}[x,y|c,e]$.
In $\mathbf{G}_{\mathrm{Cit}}$ the node $c$ appears in first place of edge $(a,c)$ and in second place of edge $(c,e)$.
Therefore, all updates for the same grounded atom are summed, which reduces the size of $\mathbf{M}$ to the size of $\mathbf{U}$.

To ensure that the updated predictions remain truth values in the range of $[0,1]$, the knowledge enhancement layer updates at first the preactivations $\mathbf{Z}$ of the GNN and then applies the activation function $\sigma$ to the updated preactivations $\mathbf{Z}'$ in order to obtain the final predictions: $\mathbf{Y}' = \sigma(\mathbf{Z}')$.
Therefore, a knowledge enhancement layer transforms $\mathbf{Z}$ to $\mathbf{Z}'$ (with $\mathbf{Z}, \mathbf{Z}' \in \mathbb{R}^{n \times m}$).
In the last step, the updates by the knowledge enhancer are added to the preactivations $\mathbf{Z}$ of the GNN and passed to $\sigma$ to obtain the updated predictions
\begin{equation*}
  \footnotesize
  \mathbf{Y'} = \sigma \Bigg( \mathbf{Z} + \sum_{\varphi \in \mathcal{K}}  \delta\mathbf{U}_{\varphi} \Bigg)
\end{equation*} 
where $\delta \mathbf{U}_{\varphi}$ is the matrix obtained by extracting the changes to the unary predicates from $\delta \mathbf{M}_{\varphi}$. 
Regarding the binary groundings, the values in $\mathbf{B}$ are set to a high positive value that results in one when $\sigma$ is applied.

\section{Related Work}
\begin{table*}[h]
  \footnotesize
  \begin{tabular}{|l|l|l|l|l|l|l|} 
  \hline
  \textbf{ Name }  & \textbf{ \#nodes } & \textbf{ \#edges } & \textbf{ \#features } & \textbf{ \#Classes } & \textbf{train/valid/test split}\\ 
  \hline
  Citeseer  & 3,327 & 9,104 & 3,703 & 6 & 1817/500/1000\\
  Cora & 2,708 & 10,556 & 1,433 & 7 & 1208/500/1000\\
  PubMed  & 19,717 & 88,648 & 500 & 3 & 18217/500/1000\\
  Flickr & 89,250 & 899,756 & 500 & 7 & 44624/22312/22312\\
  \hline
  \end{tabular}
  \centering
  \caption{
    Overview of the datasets Citeseer, Cora, PubMed and Flickr}
  \label{tab:dataset}
  \end{table*}
The field of knowledge graph completion is addressed from several research directions.
Symbolic methods exist that conduct link prediction given a set of prior knowledge
\cite{semantic_data_mining} 
\cite{anyburl}. 
Embedding-based methods \cite{electronics9050750} are mostly sub-symbolic methods to obtain node embeddings that are used for knowledge graph completion tasks.
Usually, their common objective is to find similar embeddings for nodes that are located closely in the graph. 
The majority of these methods only encodes the graph structure, but does not consider node-specific feature information \cite{nodeclassification_linkprediction}.
However, KeGNN is based on GNNs that are suited for learning representations of graphs attributed with node features. 
It stacks additional layers that interpret the outputs of the GNN in fuzzy logic and modify them to increase the satisfiability.
Therefore, it is considered a neuro-symbolic method. 
In the multifaceted neuro-symbolic field, KeGNN can be placed in the category of knowledge-guided learning \cite{kenn}, where the focus lies on learning in the presence of additional supervision introduced as prior knowledge. 
Within this category, KeGNN belongs to the model-based approaches, where prior knowledge in the form of knowledge enhancement layers is an integral part of the model \cite{relational_kenn}.
Beyond, loss-based methods such as logic tensor networks \cite{ltn} exist that encode the satisfiability of prior knowledge as an optimization objective.

Further, in \cite{delong2023neurosymbolic} neuro-symbolic approaches dealing with graph structures are classified into three categories.   
First, logically informed embedding approaches \cite{sole, 10.1007/978-3-030-88361-4_24} use predefined logical rules that provide knowledge to a neural system, while both components are mostly distinct. 
Second, approaches for knowledge graph embedding with logical constraints \cite{simple, kale} use prior knowledge as constraints on the neural knowledge graph embedding method in order to modify predictions or embeddings. 
Thirdly, neuro-symbolic methods are used for learning rules for graph reasoning tasks \cite{expressgnn, qu2021rnnlogic}. 
This allows for rule generation or confidence scores for prior knowledge and makes the models robust to exceptions or soft knowledge. 
KeGNN best falls into the second category, since the prior knowledge is interpreted in fuzzy logic to be integrated with the neural model and update the GNN's predictions.
The idea of confidence values in category three shares the common property of weighting knowledge as with KeGNN's clause weights.  
However, even though KeGNN's clause weights introduce a notion of impact of a clause when predictions are made, they cannot directly be interpreted as the confidence in a rule.

In the well-known \emph{Kautz Taxonomy} \cite{kautz} that classifies neuro-symbolic approaches according to the integration of neural and symbolic modules, KeGNN falls best into the category \texttt{Neuro[Symbolic]} (Type 6) of fully-integrated neuro-symbolic systems that embed symbolic reasoning in a neural architecture. 

\section{Experimens}
To evaluate the performance of KeGNN, we apply it to the datasets Citeseer, Cora, PubMed and Flickr that are common benchmarks for node classification in a transductive setting.
In the following, KeGNN is called KeGCN and KeGAT when instantiated to a GCN or a GAT, respectively. 
As additional baseline, we consider KeMLP, that stacks knowledge enhancement layers onto an MLP, as proposed in \cite{relational_kenn}.
Further, the standalone neural models MLP, GCN and GAT are used as baselines. 
While Citeseer, Cora and PubMed are citation graphs that encode citations between scientific papers (as in Example \ref{ex:prior_knowledge_cit}), Flickr contains images and shared properties between them. 
All datasets can be modelled as homogeneous, labelled and attributed graphs as defined in Section~\ref{sec:graphstructured_data}.
Table~\ref{tab:dataset} gives an overview of the named datasets in this work. 
The datasets are publicly available on the dataset collection\footnote{\url{https://pytorch-geometric.readthedocs.io/en/latest/modules/datasets.html}} of PyTorch Geometric \cite{pytorch_geometric}.
For the split into train, valid and test set, we take the predefined splits in \cite{fast_gcn} for the citation graphs and in \cite{graph_saint} for Flickr. 
Word2Vec vectors \cite{word2vec} are used as node features for the citation graphs and image data for Flickr.
Figure~\ref{fig:graph_structure} visualizes the graph structure of the underlying datasets in this work as a homogeneous, attributed and labelled graph on the example of Citeseer. 

The set of prior logic for the knowledge enhancement layers is manually defined. 
In this work, we encode the assumption that the existence of an edge for a node pair points to their membership to the same class and hence provides added value to the node classification task. 
In the context of citation graphs, this implies that two documents that cite each other refer to the same topic, while for Flickr, linked images share the same properties. 
Following this pattern for all datasets, a clause $\mathrm{\varphi}$:
$\forall xy: \neg \mathrm{Cls_i(x)} \lor$ $\neg \mathrm{Link(x,y)} \lor \mathrm{Cls_i(y)}$ is instantiated for each node class $\mathrm{Cls_i}, \mathrm{i \in \{1, \hdots, m\}}$.
More details on the experiments are given in Section~\ref{sec:experiment_details}. 
The source code of the experiments are publicly available\footnote{\url{https://gitlab.inria.fr/tyrex/kegnn}\label{gitlab}}.
.
\subsection{Results}
   
\begin{table*}
  \footnotesize
  \centering
  \begin{tabular}{|l|ll|ll|ll|} 
  \hline
   & \textbf{MLP} & \textbf{KeMLP} & \textbf{GCN} & \textbf{KeGCN} & \textbf{GAT} & \textbf{KeGAT} \\ 
  \hline
  \textbf{Cora} & \begin{tabular}[c]{@{}l@{}}0.7098 \\(0.0080)\end{tabular} & \begin{tabular}[c]{@{}l@{}}0.8072 \\(0.0193)\end{tabular} & \begin{tabular}[c]{@{}l@{}}0.8538 \\(0.0057)\end{tabular} & \begin{tabular}[c]{@{}l@{}}\textbf{0.8587 }\\(0.0057)\end{tabular} & \begin{tabular}[c]{@{}l@{}}0.8517 \\(0.0068)\end{tabular} & \begin{tabular}[c]{@{}l@{}}0.8498 \\(0.0066)\end{tabular} \\ 
  \hline
  \textbf{CiteSeer} & \begin{tabular}[c]{@{}l@{}}0.7278 \\(0.0081)\end{tabular} & \begin{tabular}[c]{@{}l@{}}0.7529 \\(0.0067)\end{tabular} & \begin{tabular}[c]{@{}l@{}}0.748 \\(0.0102)\end{tabular} & \begin{tabular}[c]{@{}l@{}}0.7506 \\(0.0096)\end{tabular} & \begin{tabular}[c]{@{}l@{}}0.7718 \\(0.0072)\end{tabular} & \begin{tabular}[c]{@{}l@{}}\textbf{0.7734 }\\(0.0073)\end{tabular} \\ 
  \hline
  \textbf{PubMed} & \begin{tabular}[c]{@{}l@{}}0.8844 \\(0.0057)\end{tabular} & \begin{tabular}[c]{@{}l@{}}\textbf{0.8931} \\(0.0048)\end{tabular} & \begin{tabular}[c]{@{}l@{}}0.8855 \\(0.0062)\end{tabular} & \begin{tabular}[c]{@{}l@{}}0.8840 \\(0.0087)\end{tabular} & \begin{tabular}[c]{@{}l@{}}0.8769 \\(0.0040)\end{tabular} & \begin{tabular}[c]{@{}l@{}}0.8686 \\(0.0081)\end{tabular} \\ 
  \hline
  \textbf{Flickr} & \begin{tabular}[c]{@{}l@{}}0.4656 \\(0.0018)\end{tabular} & \begin{tabular}[c]{@{}l@{}}0.4659 \\(0.0012)\end{tabular} & \begin{tabular}[c]{@{}l@{}}\textbf{0.5007 }\\(0.0063)\end{tabular} & \begin{tabular}[c]{@{}l@{}}0.4974 \\(0.0180)\end{tabular} & \begin{tabular}[c]{@{}l@{}}0.4970\\(0.0124)\end{tabular} & \begin{tabular}[c]{@{}l@{}}0.4920\\(0.0189)\end{tabular} \\
  \hline
  \end{tabular}
  \caption{
      Average test accuracy of 50 runs (10 for Flickr). The standard deviations are reported in brackets. 
    }
  \label{tab:results}
  \end{table*}
To compare the performance of all models, we examine the average test accuracy over 50 runs (10 for Flickr) for the knowledge enhanced models KeMLP, KeGCN, KeGAT and the standalone base models MLP, GCN, GAT on the named datasets. 
The results are given in Table~\ref{tab:results}.
For Cora and Citeseer, KeMLP leads to a significant improvement over MLP (p-value of one-sided t-test $\ll 0.05$). 
In contrast, no significant advantage of KeGCN or KeGAT in comparison to the standalone base model is observed. 
Nevertheless, all GNN-based models are significantly superior to KeMLP for Cora. 
This includes not only KeGCN and KeGAT, but also the GNN baselines.
For Citeseer, KeGAT and GAT both outperform KeMLP.
In the case of PubMed, only a significant improvement of KeMLP over MLP can be observed, while the GNN-based models and their enhanced versions do not provide any positive effect. 
For Flickr, no significant improvement between the base model and the respective knowledge enhanced model can be observed. 
Nevertheless, all GNN-based models outperform KeMLP, reporting significantly higher mean test accuracies for KeGAT, GAT, GCN and KeGCN.


\subsubsection{Exploitation of the Graph Structure}
\begin{figure}
  \centering
  \begin{subfigure}[b]{0.5\textwidth}
    \centering
     \includegraphics[width=7cm]{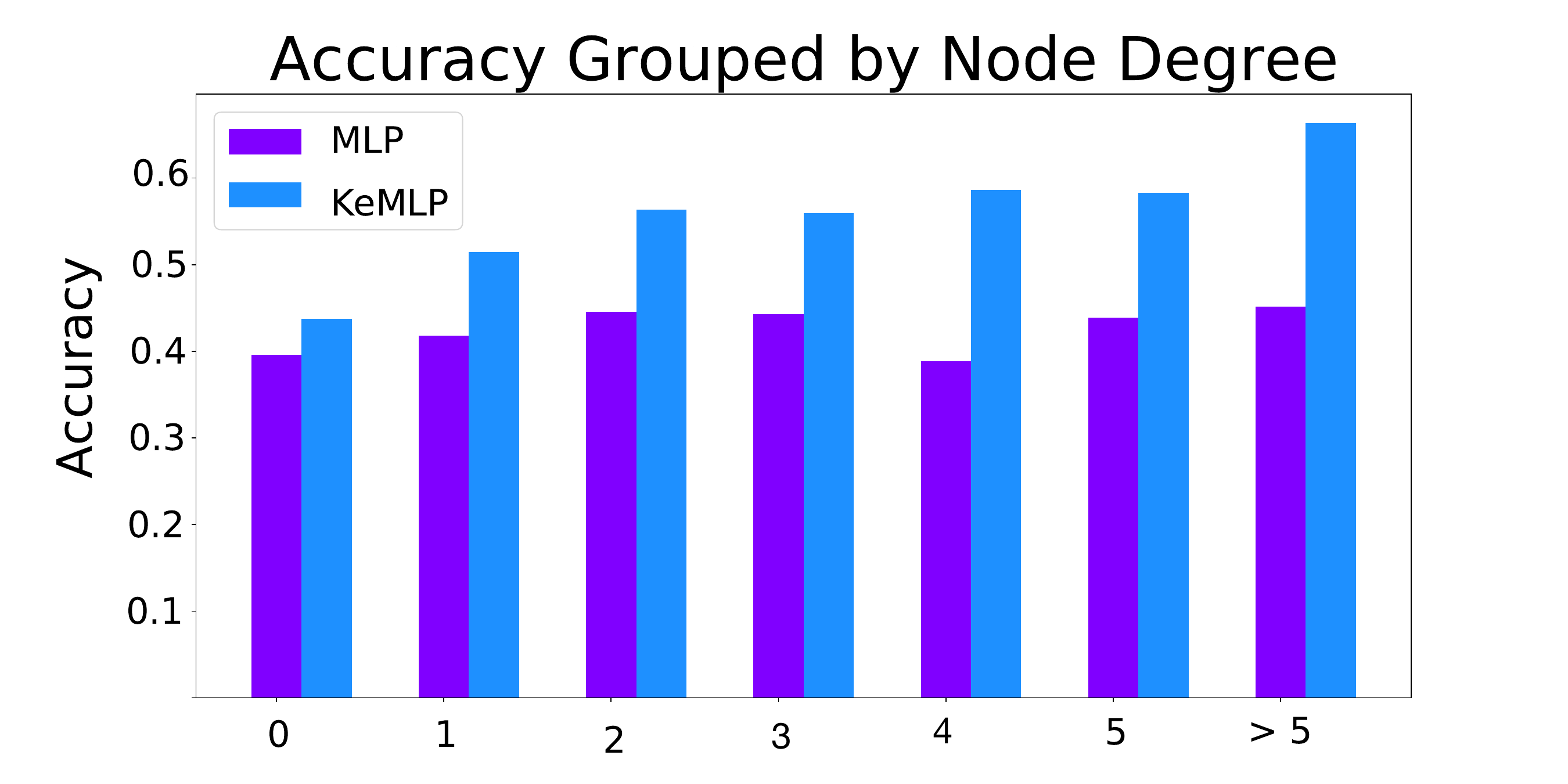}
  \end{subfigure}  
  \begin{subfigure}[b]{0.5\textwidth}
     \centering
     \includegraphics[width=7cm]{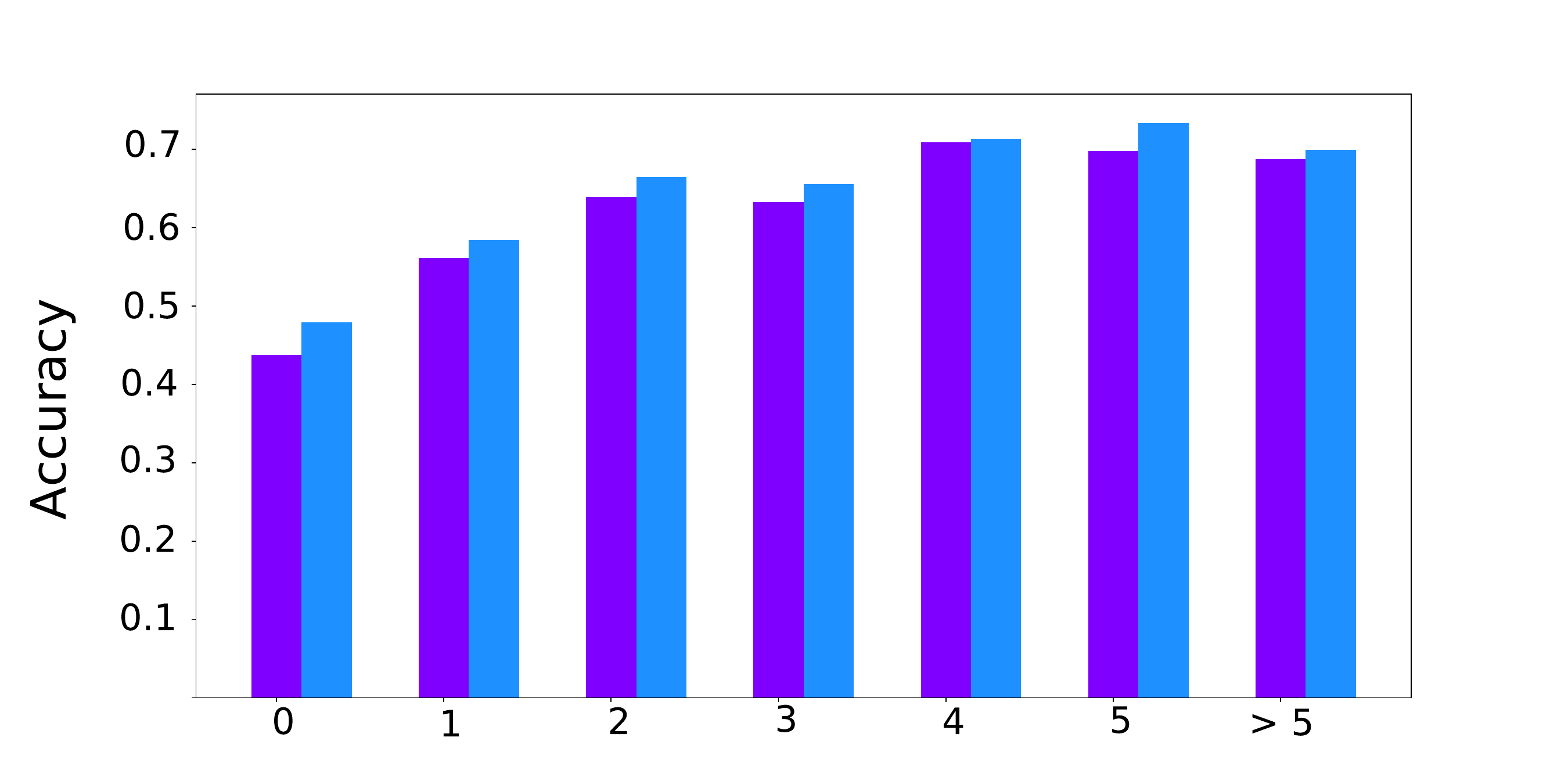}
  \end{subfigure}
\begin{subfigure}[b]{0.5\textwidth}
  \centering
  \includegraphics[width=7cm]{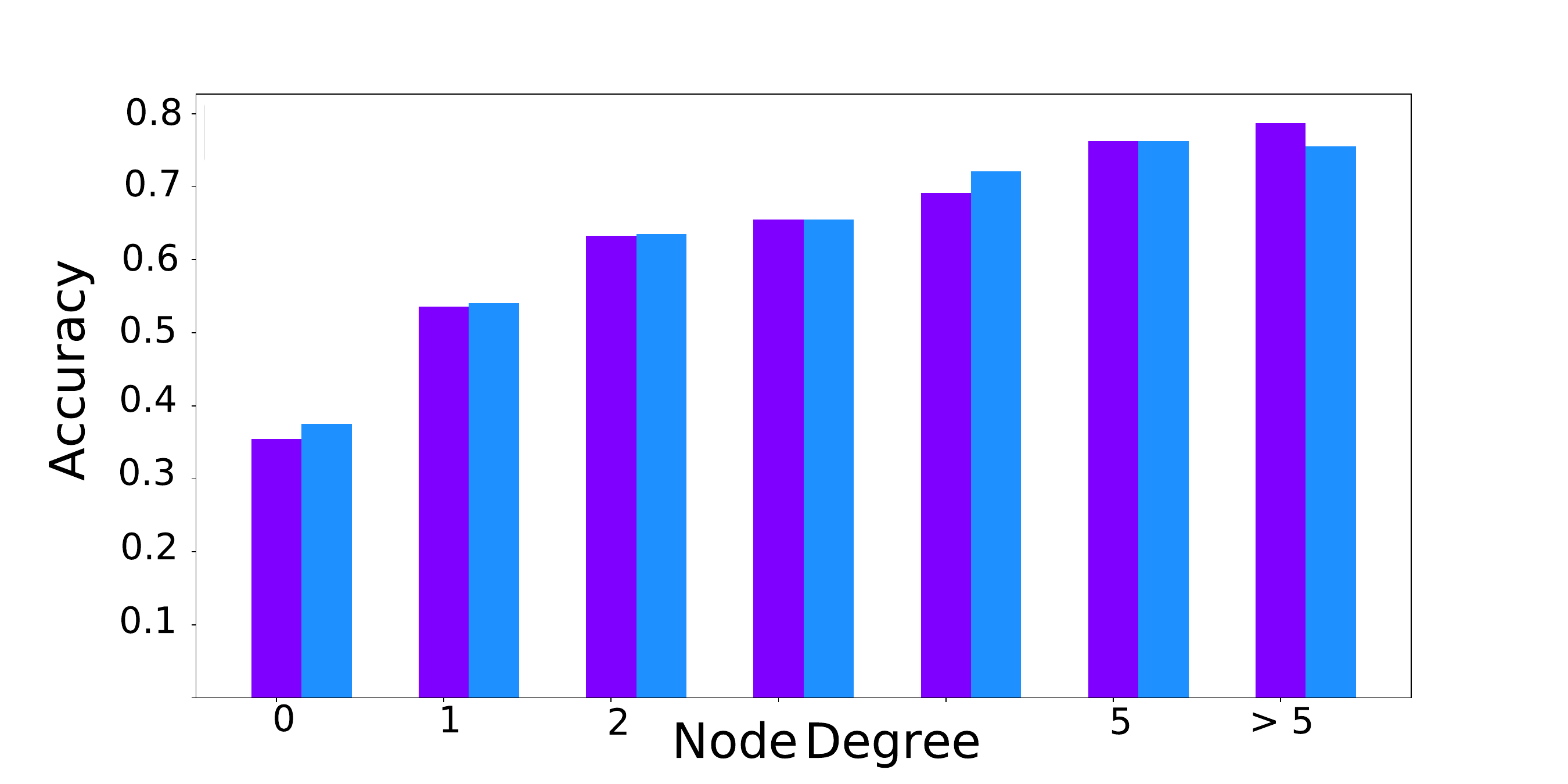} 
  \end{subfigure}
  \caption{
      The accuracy grouped by the node degree for MLP vs. KeMLP (above) and GCN vs. KeGCN (middle) and GAT vs. KeGAT(below) on Citeseer.
  \label{fig:acc_nodedegree}} 
  \end{figure}
It turns out that the performance gap between MLP and KeMLP is larger than for KeGNN in comparison to the standalone GNN.
To explain this observation, we examine how the graph structure affects the prediction performance. 
Therefore, in Figure~\ref{fig:acc_nodedegree} we analyze the accuracy grouped by the node degree for the entire graph for MLP vs. KeMLP and GCN vs. KeGCN. 
The findings for KeGAT are in line with those for KeGCN.
It is observed that KeMLP performs better compared to MLP as the node degree increases.
By contrast, when comparing GCN and KeGCN, for both models, the accuracy is observed superior for nodes with a higher degree.

This shows that rich graph structure is helpful for the node classification in general. 
Indeed, the MLP is a simple model that misses information on the graph structure and thus benefits from graph structure contributed by KeMLP in the form of binary predicates.
On the contrary, standalone GNNs can process graph structure by using message passing techniques to transmit learned node representations between neighbors. 
The prior knowledge introduced in the knowledge enhancer is simple.
It encodes that two neighbors are likely to be of the same class. 
An explanation for the small difference in performance is that GNNs may be able to capture and propagate this simple knowledge across neighbors implicitly, using its message passing technique.  
In other words we observe that, in this particular case,  the introduced knowledge happens to be redundant for GNNs. However, the introduced knowledge significantly improves the accuracy of MLPs. 
In this context, we discuss 
perspectives for future work in Section~\ref{sec:limits}.
%
%

\subsubsection{Robustness to wrong knowledge}
Furthermore, a question of interest is how the knowledge enhanced model find a balance between knowledge and graph data in case of knowledge that is not consistent with the graph data. 
In other words, can the KeGNN successfully deal with nodes having mainly neighbors that belong to a different ground truth class and thus contribute misleading information to the node classification? 

\begin{figure}[t]
  \centering
  \begin{subfigure}{4cm}
  \centering\includegraphics[width=4cm]{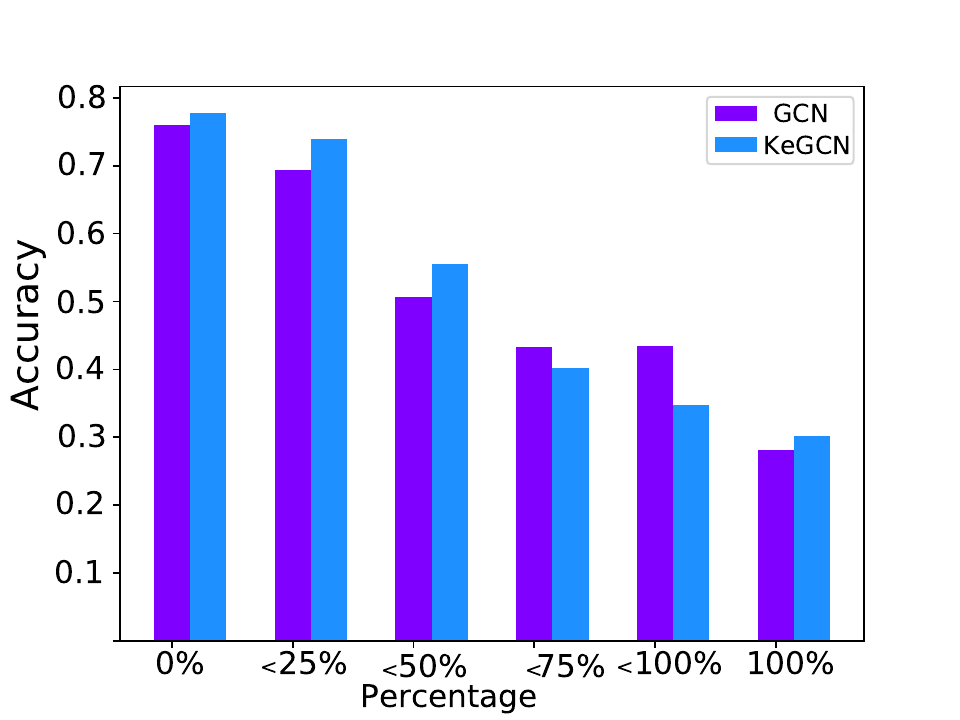}
  \end{subfigure}%
  \begin{subfigure}{4cm}
  \centering\includegraphics[width=4cm]{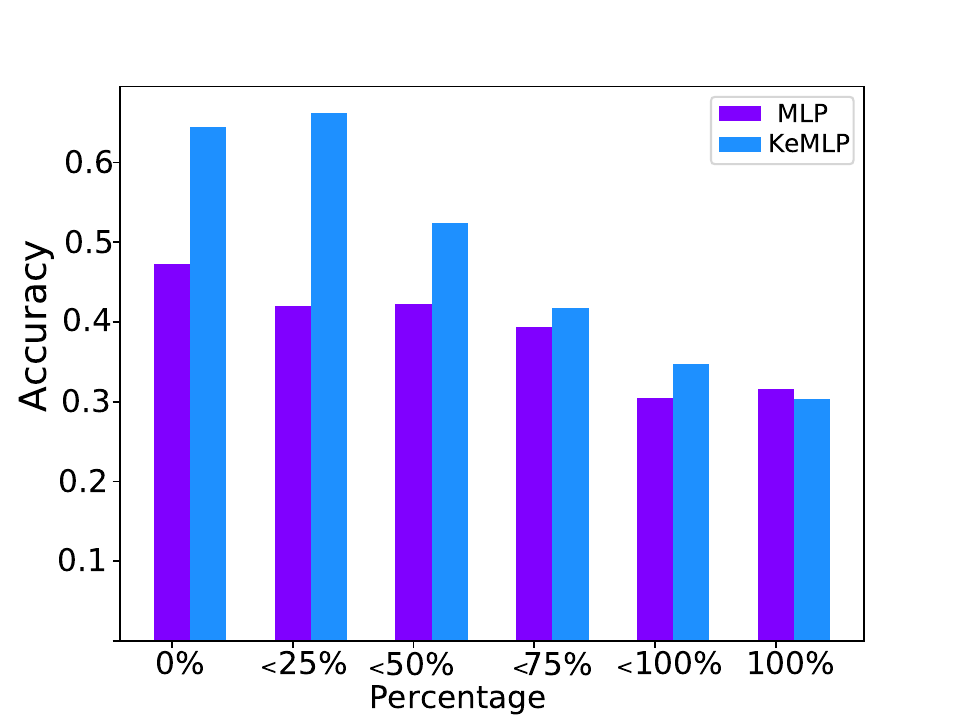}
  \end{subfigure}\vspace{10pt}
  \begin{subfigure}{4cm}
    \centering\includegraphics[width=4cm]{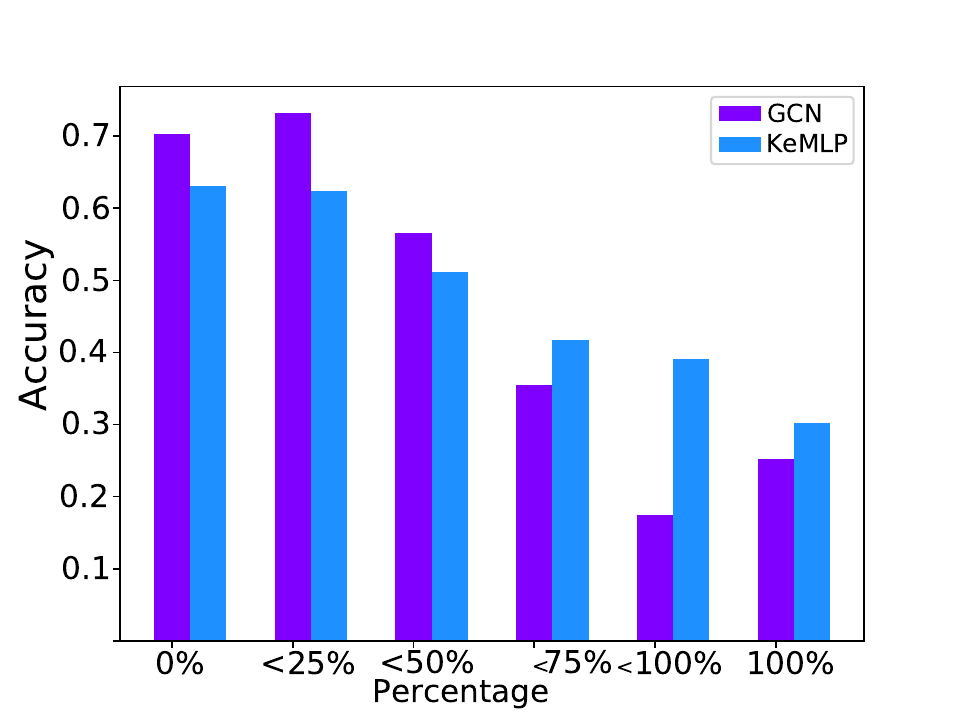}
    \end{subfigure}\vspace{10pt}
  \caption{
      The accuracy grouped by the ratio of misleading first-order neighbors for GCN vs. KeGCN (left), MLP vs. KeMLP (right), GCN vs. KeMLP (below) on Citeseer. 
      \label{fig:acc_wrong_neighbors}} 
  \end{figure}

To analyze this question, we categorize the accuracy by the proportion of misleading nodes in the neighborhood, see Figure~\ref{fig:acc_wrong_neighbors}.
Misleading nodes are node neighbors that have a different ground truth class than the node to be classified. 
It turns out that KeMLP is particularly helpful over MLP when the neighborhood provides the right information.
However, if the neighborhood is misleading (if most or even all of the neighbors belong to a different class), an MLP that ignores the graph structure can lead to even better results.
When comparing KeGCN and GCN, there is no clear difference. 
This is expected, since both models are equally affected by misleading nodes as they utilize the graph structure. 
Just as a GCN, the KeGCN is not necessarily robust to wrong prior knowledge since the GCN component uses the entire neighborhood, including the misleading nodes.

When comparing GCN to KeMLP, see plot below in Figure~\ref{fig:acc_wrong_neighbors}, KeMLP is more robust to misleading neighbors. 
While GCN takes the graph structure as given and includes all neighbors equally in the embeddings by graph convolution, the clause weights in the knowledge enhancement layers provide a way to devalue knowledge.
If the data frequently contradicts a clause, the model has the capacity to reduce the respective clause weight in the learning process and reduce its impact.  

\subsubsection{Clause Weight Learning}
Further, we want to examine whether the clause weights learned during training are aligned with the knowledge in the ground truth data.
The clause weights provide insights on the magnitude of the updates made by a clause. 
\begin{figure}[h]
  \centering
  \begin{subfigure}{4cm}
  \centering\includegraphics[width=4cm]{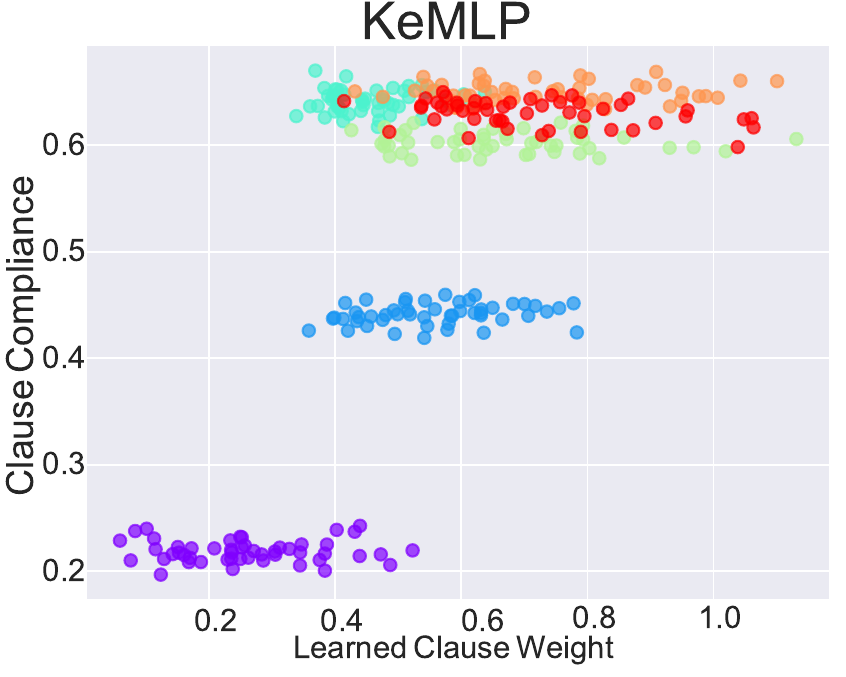}
  \end{subfigure}%
  \begin{subfigure}{4cm}
  \centering\includegraphics[width=4cm]{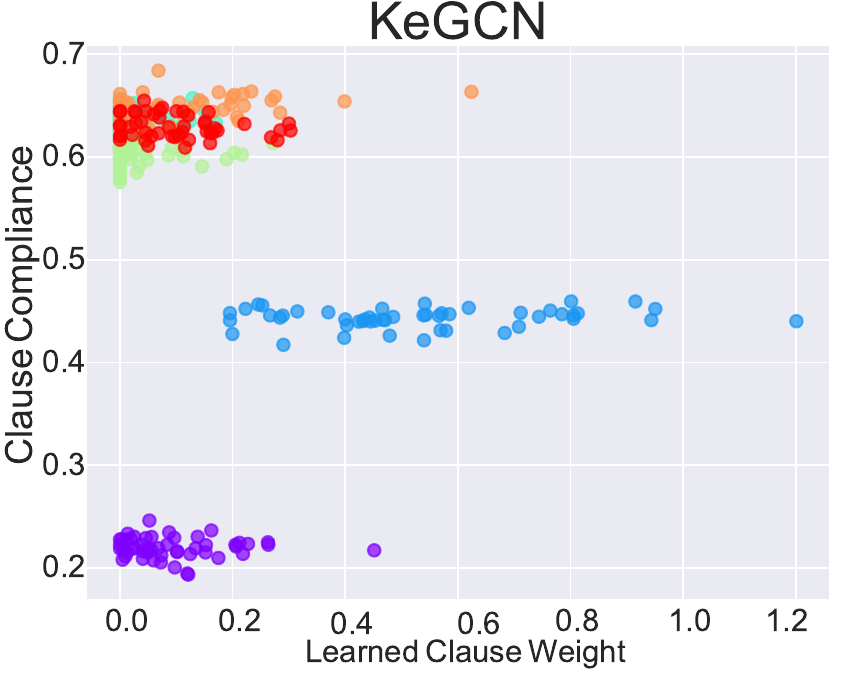}
  \end{subfigure}\vspace{10pt} 
  \caption{
      Learned clause weights vs. clause compliance for KeMLP (left) and KeGCN (right) on Citeseer.
    \label{fig:clause_weights}} 
  \end{figure}
The \emph{clause compliance} \cite{kenn} measures how well the prior knowledge is satisfied in a graph. 
Given a clause $\varphi$, a class $\mathrm{Cls_i}$, a set of nodes $\mathbf{V}$, a set of nodes of the class $\mathrm{Cls_i}$: $\mathbf{V_i} = \{v_i | v_i \in \mathbf{V} \land \mathrm{Cls}(v_i) == \mathrm{i}\}$, and the neighborhood $\mathcal{N}(v_i)$ of $v_i$, the clause compliance of clause $\varphi$ on graph $\mathbf{G}$ is defined as follows:     
\begin{equation}
  \label{eq:compliance}
  \operatorname{Compliance}(\mathbf{G}, \varphi) = \frac{\sum_{v_i \in \mathbf{V_i}} \sum_{v_j \in \mathcal{N}(v)} \mathbf{1} [\text{ if } v_j \in \mathbf{V_i}]}{\sum_{v_i \in \mathbf{V_i}} | \mathcal{N}(v_i)|}
\end{equation}
\label{sec:clause_weight_evaluation_appendix}
In other words, the clause compliance counts how often among nodes of a class $\mathrm{Cls_i}$ the neighboring nodes have the same class \cite{kenn}.
The clause compliance can be calculated on the ground truth classes of the training set or the predicted classes. 
As a reference, we measure the clause compliance based on the ground truth labels in the training set. 
Figure~\ref{fig:clause_weights} displays the learned clause weights for KeGCN and KeMLP versus the clause compliance on the ground truth labels of the training set.
For KeMLP, a positive correlation between the learned clause weights and the clause compliance on the training set is observed. 
This indicates that higher clause weights are learned for clauses that are satisfied in the training set. 
Consequently, these clauses have a higher impact on the updates of the predictions. 
In addition, the clause weights corresponding to clauses with low compliance values make smaller updates to the initial predictions. 
Accordingly, clauses that are rarely satisfied learn lower clause weights during the training process.
In the case of KeGCN, the clause weights are predominantly set to values close to zero. 
This is in accordance with the absence of a significant performance gap between GCN and KeGCN. 
Since the GCN itself already leads to valid classifications, smaller updates are required by the clause enhancers. 
\begin{figure}[]
  \centering
  \begin{subfigure}{4cm}
  \centering\includegraphics[width=4cm]{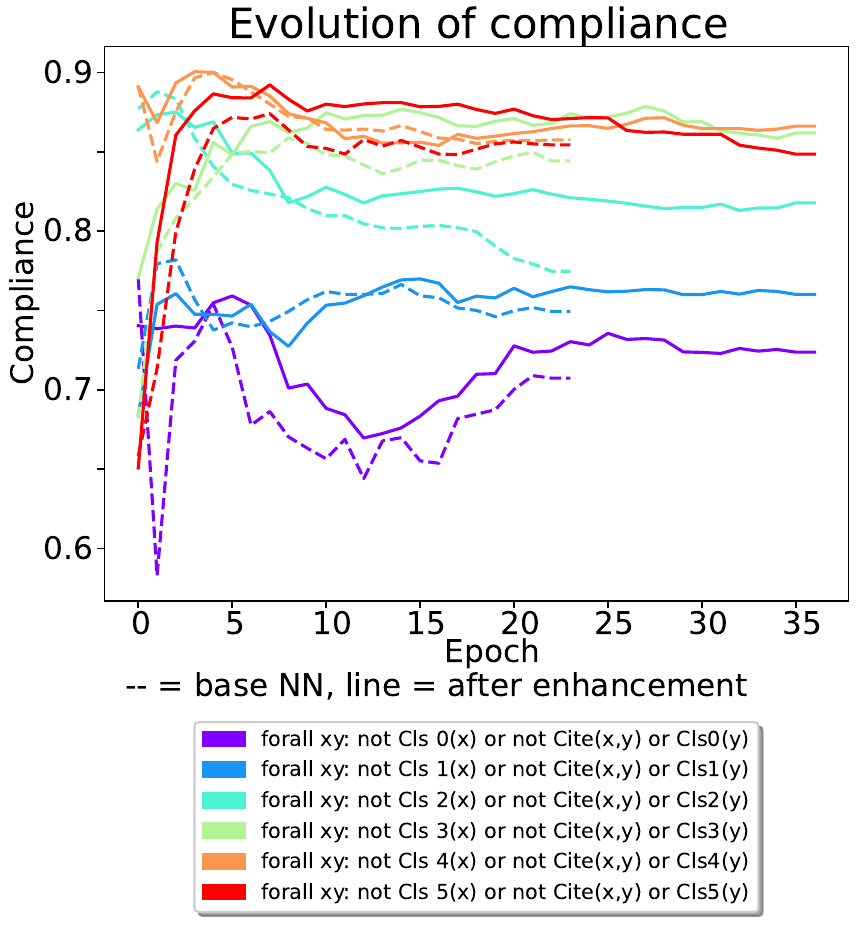}
  \end{subfigure}%
  \begin{subfigure}{4cm}
  \centering\includegraphics[width=4cm]{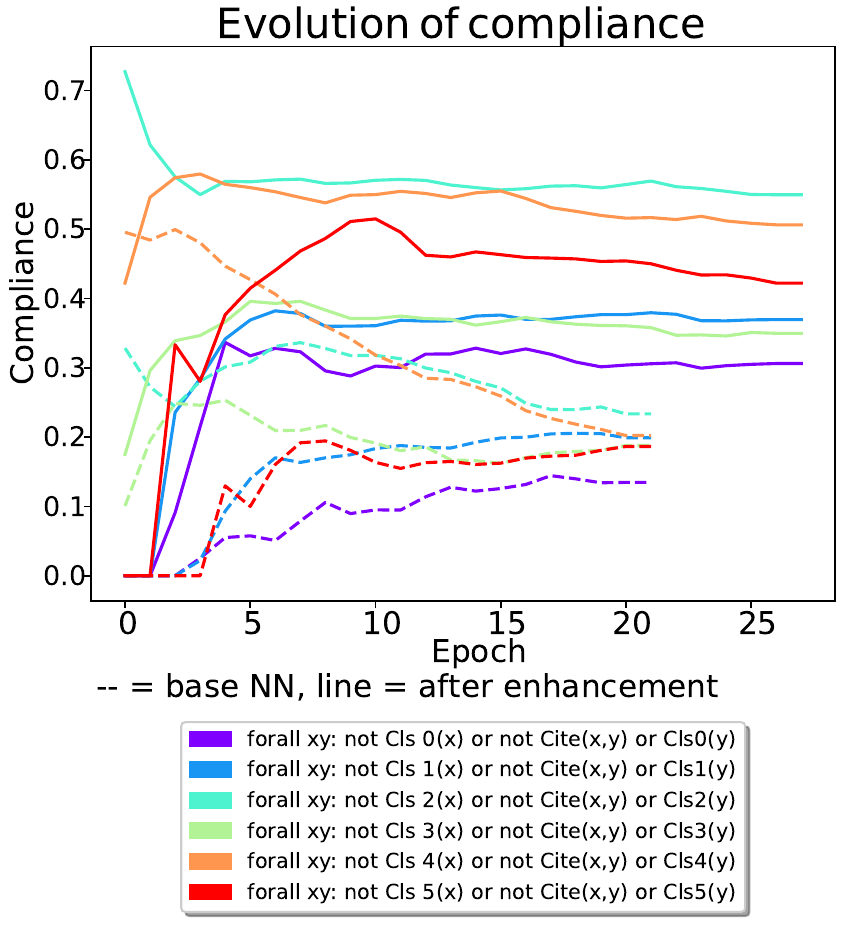}
  \end{subfigure}\vspace{10pt}
  \caption{
      Clause compliance during training for GCN vs. KeGCN (left) and MLP vs. KeMLP (right) on Citeseer. 
  \label{fig:compliance_evolution}} 
  \end{figure}

  Furthermore, we analyze how the compliance evolves during training to investigate whether the models learn predictions that increase the satisfaction of the prior knowledge. 
Figure~\ref{fig:compliance_evolution} plots the evolution of the clause compliance for the six clauses for GCN vs. KeGCN and MLP vs. KeMLP.
It is observed that GCN and KeGCN yield similar results as the evolution of the compliance during training for both models is mostly aligned. 
For MLP vs. KeMLP the clause compliance of the prediction of the MLP converges to lower values for all classes than the clause compliance obtained with the KeMLP. 
This gives evidence that the knowledge enhancement layer actually improves the satisfiability of the prior knowledge. 
As already observed, this gives evidence that the standalone GCN is able to implicitly satisfy the prior knowledge even though it is not explicitly defined.

\subsection{Additional Experiment Details}
\label{sec:experiment_details}

\subsubsection{Implementation}
The code\footnotemark[4] is based on PyTorch \cite{pytorch} and the graph learning library PyTorch Geometric \cite{pytorch_geometric}. 
The Weights \& Biases tracking tool \cite{wandb} is used to monitor the experiments.
All experiments are conducted on a machine running an Ubuntu 20.4 equipped with an Intel(R) Xeon(R) Silver 4114 CPU 2.20GHz processor, 192G of RAM and
one GPU Nvidia Quadro P5000.

\subsubsection{Model Parameters and Hyperparameter Tuning}
\label{sec:hyperparameter_tuning}
KeGNN contains a set of hyperparameters.
Batch normalization \cite{batch_norm} is applied after each hidden layer of the GNN.
The Adam optimizer \cite{adam} is used as optimizer for all models. 
Concerning the hyperparameters specific to the knowledge enhancement layers, the initialization of the preactivations of the binary predicates (which are assumed to be known) is taken as a hyperparameter. 
They are set to a high positive value for edges that are known to exist and correspond to the grounding of the binary predicate.
Furthermore, different initializations of clause weights and constraints on them are tested. 
Moreover, the number of stacked knowledge enhancement layers is a hyperparameter.
We further allow the model to randomly neglect a proportion of edges by setting an edges drop rate parameter.
Further, we test whether the normalization of the edges with the diagonal matrix 
$\tilde {\mathbf{D}} = \sum_{j} \tilde{\mathbf{A}}_{i,j}$ 
(with $\tilde{\mathbf{A}} = \mathbf{A} + \mathbf{I}$) is helpful. 

To find a suitable set hyperparameters for each dataset and model, we perform a random search with up to 800 runs and 48h time limit and choose the parameter combination which leads to the highest accuracy on the validation set.
The hyperparameter tuning is executed in Weights and Biases \cite{wandb}.
The following hyperparameter values are tested:
\begin{itemize}
  \item Adam optimizer parameters: $\beta_1$: 0.9, $\beta_2$: 0.99, $\epsilon$: 1e-07
  \item Attention heads: $\{1, 2, 3, 4, 6, 8, 10\}$
  \item Batch size: $\{128, 512, 1024, 2048, \text{full batch}\}$
  \item Binary preactivation: $\{0.5, 1.0, 10.0, 100.0, 500.0\}$
  \item Clause weights initialization: $\{0.001, 0.1, 0.25, 0.5$, random uniform distribution on [0,1)\}
  \item Dropout rate: $0.5$
  \item Edges drop rate: random uniform distribution $[0.0, 0.9]$
  \item Edge normalization: $\{\text{true, false}\}$
  \item Early stopping: $\delta_{min}: 0.001$, patience: \{1, 10, 100\}
  \item Hidden layer dimension: \{32, 64, 128, 256\}
  \item Learning rate: random uniform distribution $[0.0001, 0.1]$
  \item Clause weight clipping: $w_{min}: 0.0$, $w_{max}$: random uniform distribution: $[0.8, 500.0]$
  \item Number of knowledge enhancement layers: \\$\{1, 2, 3, 4, 5, 6\}$
  \item Number of hidden layers: $\{2, 3, 4, 5, 6\}$
  \item Number of epochs $200$ (unless training stopped early)

\end{itemize}
The obtained parameter combinations for the models KeMLP, KeGCN and KeGAT for Cora, Citeseer, PubMed and Flickr are displayed 
in Table~\ref{tab:pubmed_flickr_hp}.
We set the random seed for all experiments to 1234. 

The reference models MLP, GCN and GAT are trained with the same parameter set as the respective knowledge enhanced models. 

\section{Limitations and Perspectives}
\begin{table*}[h]
  \footnotesize
  \centering
  \begin{tblr}{
    cell{2}{1} = {r=3}{c},
    cell{5}{1} = {r=3}{c},
    cell{8}{1} = {r=3}{c},
    cell{11}{1} = {r=3}{c},
    vlines,
    hline{1-2,5,8,11,14} = {-}{},
    hline{3-4,6-7,9-10,12-13} = {2-13}{},
  }
   & \textbf{mode} & {\textbf{atten}\\\textbf{-tion }\\\textbf{heads}} & {\textbf{batch }\\\textbf{size~}} & {\textbf{binary }\\\textbf{preacti-}\\\textbf{vation}} & {\textbf{initial }\\\textbf{clause }\\\textbf{weight}} & {\textbf{edges }\\\textbf{drop }\\\textbf{rate}} & {\textbf{es }\\\textbf{pati-}\\\textbf{ence}} & {\textbf{hidden }\\\textbf{chan-}\\\textbf{nels}} & {\textbf{learn-}\\\textbf{ing }\\\textbf{rate~}} & {\textbf{norma}\\\textbf{-lize }\\\textbf{edges}} & {\textbf{KE }\\\textbf{layers}} & {\textbf{hidden }\\\textbf{layers}}\\
  \textbf{PubMed} & \textbf{KeMLP} & - & 1024 & 10.0 & 0.001 & 0.22 & 100 & 256 & 0.057 & false & 2 & 4\\
   & \textbf{KeGCN} & - & full batch & 1.0 & random & 0.66 & 10 & 256 & 0.043 & false & 1 & 2\\
   & \textbf{KeGAT} & 8 & 1024 & 10.0 & 0.5 & 0.07 & 10 & 256 & 0.016 & true & 5 & 2\\
  \textbf{Flickr} & \textbf{KeMLP} & - & 128 & 10.0 & 0.001 & 0.2 & 10 & 32 & 0.001 & true & 1 & 2\\
   & \textbf{KeGCN} & - & 1024 & 500.0 & 0.001 & 0.24 & 10 & 128 & 0.016 & true & 4 & 4\\
   & \textbf{KeGAT} & 8 & 2048 & 500.0 & 0.1 & 0.12 & 100 & 64 & 0.0039 & false & 1 & 3\\
  \textbf{Cora} & \textbf{\textbf{KeMLP}} & - & 512 & 10.0 & 0.5 & 0.47 & 1 & 32 & 0.026 & true & 4 & 2\\
   & \textbf{\textbf{KeGCN}} & - & 512 & 100.0 & random & 0.17 & 1 & 256 & 0.032 & false & 2 & 2\\
   & \textbf{\textbf{KeGAT}} & 1 & full batch & 1.0 & 0.5 & 0.27 & 10 & 64 & 0.033 & true & 1 & 2\\
  \textbf{Citeseer} & \textbf{\textbf{\textbf{\textbf{KeMLP}}}} & - & 128 & 10.0 & 0.5 & 0.01 & 10 & 256 & 0.028 & true & 1 & 2\\
   & \textbf{\textbf{\textbf{\textbf{KeGCN}}}} & - & full batch & 0.5 & 0.25 & 0.35 & 10 & 128 & 0.037 & false & 3 & 5\\
   & \textbf{\textbf{\textbf{\textbf{KeGAT}}}} & 3 & 1024 & 0.5 & 0.1 & 0.88 & 10 & 32 & 0.006 & true & 2 & 2
  \end{tblr}
  \caption{
      Hyperparameters and experiment configuration for PubMed and Flickr
    }
  \label{tab:pubmed_flickr_hp}
  \end{table*}
\label{sec:limits}
The method of KeGNN is limited in some aspects, which we present in this section. 
In this work, we focus on homogeneous graphs.
In reality, however, graphs are often heterogeneous with multiple node and edge types \cite{heterogeneous_gnn}.
Adaptations are necessary on both the neural and the symbolic side to apply KeGNN to heterogeneous graphs. 
The restriction to homogeneous graphs also limits the scope of formulating complex prior knowledge.
Eventually, the datasets used in this work and the set of prior knowledge are too simple for KeGNN to exploit its potential and lead to a significant improvement over the GNN.
The experimental results show that the knowledge encoded in the symbolic component leads to significant improvement over an MLP that is not capable to capture and learn that knowledge. 
This indicates that for more complex knowledge that is harder for a GNN to learn, KeGNN has the potential to bring higher improvements. 
A perspective for further work is the extension of KeGNN to more generic data structures such as incomplete and heterogeneous knowledge graphs in conjunction with more complex prior knowledge.

Another limitation of KeGNN is scalability. 
With an increasing number of stacked knowledge enhancement layers, the affected node neighborhood grows exponentially, which can lead to significant memory overhead.
This problem is referred as {neighborhood explosion} \cite{gnn_kg} and is particularly problematic in the context of training on memory-constrained GPUs. 
This affects both the GNN and the knowledge enhancement layers that encode binary knowledge.
Methods from scalable graph learning \cite{gnn_autoscale, graph_saint, graph_sage} represent potential solutions for the neighborhood explosion problem in KeGNN. 

Furthermore, limitations appear in the context of link prediction with KeGNN.
For link prediction, a neural component is required that predicts fuzzy truth values for binary predicates.   
At present, KeGNN can handle clauses containing binary predicates, but their truth values are initialized with artificial predictions, where a high value encodes the presence of an edge.
This limits the application of KeGNN to datasets for which the graph structure is complete and known a priori.

\section{Conclusion}
In this work, we introduced KeGNN, a neuro-symbolic model that integrates GNNs with symbolic knowledge enhancement layers to create an end-to-end differentiable model. 
This allows the use of prior knowledge to improve node classification while exploiting the strength of a GNN to learn expressive representations.
Experimental studies show that the inclusion of prior knowledge has the potential to improve simple neural models (as observed in the case of MLP).
However, the knowledge enhancement of GNNs is harder to achieve on the underlying and limited benchmarks for which the injection of simple knowledge concerning local neighborhood is redundant with the representations that GNNs are able to learn.
Nevertheless, KeGNN has not only the potential to improve graph completion tasks from a performance perspective, but also to increase interpretability through clause weights. 
This work is a step towards a holistic neuro-symbolic method on incomplete and noisy semantic data, such as knowledge graphs. 
\section*{Acknowledgments}
This work has been partially supported by the MIAI Knowledge communication and evolution chair (ANR-19-P3IA-0003). 

\bibliographystyle{IEEEtran}
\bibliography{sample-ceur}
\end{document}